\pdfoutput=1

\documentclass[11pt]{article}

\usepackage{naacl2021}

\usepackage{times}
\usepackage{latexsym}
\usepackage{amsmath, amssymb}
\usepackage{multirow}
\usepackage{graphicx}
\graphicspath{ {./} }

\usepackage[T1]{fontenc}

\usepackage[utf8]{inputenc}

\usepackage{microtype}

\title{Disentangling Semantics and Syntax in Sentence Embeddings \\ with Pre-trained Language Models}

{
\centering
\author{James Y. Huang \and Kuan-Hao Huang \and Kai-Wei Chang \\
University of California, Los Angeles \\
\texttt{\{jyhuang, khhuang, kwchang\}@cs.ucla.edu}}
}

\begin{document}
\maketitle

\begin{abstract}
Pre-trained language models have achieved huge success on a wide range of NLP tasks. However, contextual representations from pre-trained models contain entangled semantic and syntactic information, and therefore cannot be directly used to derive useful semantic sentence embeddings for some tasks. Paraphrase pairs offer an effective way of learning the distinction between semantics and syntax, as they naturally share semantics and often vary in syntax. In this work, we present ParaBART, a semantic sentence embedding model that learns to disentangle semantics and syntax in sentence embeddings obtained by pre-trained language models. ParaBART is trained to perform syntax-guided paraphrasing, based on a source sentence that shares semantics with the target paraphrase, and a parse tree that specifies the target syntax. In this way, ParaBART learns disentangled semantic and syntactic representations from their respective inputs with separate encoders. Experiments in English show that ParaBART outperforms state-of-the-art sentence embedding models on unsupervised semantic similarity tasks. Additionally, we show that our approach can effectively remove syntactic information from semantic sentence embeddings, leading to better robustness against syntactic variation on downstream semantic tasks. 
\end{abstract}

\section{Introduction}

Semantic sentence embedding models encode sentences into fixed-length vectors based on their semantic relatedness with each other. If two sentences are more semantically related, their corresponding sentence embeddings are closer.
As sentence embeddings can be used to measures semantic relatedness without requiring supervised data, they have been used in many applications, such as semantic textual similarity \citep{agirre2016semeval-2016}, question answering \citep{nakov2017semeval-2017}, and natural language inference \citep{artetxe2019massively}. 

Recent years have seen huge success of pre-trained language models across a wide range of NLP tasks (\citealp{devlin2019bert}; \citealp{lewis2020bart}). However, several studies (\citealp{reimers2019sentence-bert}; \citealp{li-etal-2020-sentence}) have found that sentence embeddings from pre-trained language models perform poorly on semantic similarity tasks when the models are not fine-tuned on task-specific data. Meanwhile, \citet{DBLP:journals/corr/abs-1901-05287} shows that BERT without fine-tuning performs surprisingly well on syntactic tasks. Hence, we posit that these contextual representations from pre-trained language models without fine-tuning capture entangled semantic and syntactic information, and therefore are not suitable for sentence-level semantic tasks. 

Ideally, the semantic embedding of a sentence should not encode its syntax, and two semantically similar sentences should have close semantic embeddings regardless of their syntactic differences. While various models (\citealp{conneau-etal-2017-supervised}; \citealp{cer2018universal}; \citealp{reimers2019sentence-bert})  have been proposed to improve the performance of sentence embeddings on downstream semantic tasks, most of these approaches do not attempt to separate syntactic information from sentence embeddings.  

To this end, we propose ParaBART, a semantic sentence embedding model that learns to disentangle semantics and syntax in sentence embeddings.
Our model is built upon BART \citep{lewis2020bart}, a sequence-to-sequence Transformer \citep{NIPS2017_3f5ee243} model pre-trained with self-denoising objectives. Parallel paraphrase data is a good source of learning the distinction between semantics and syntax, as paraphrase pairs naturally share the same meaning but often differ in syntax. Taking advantage of this fact, ParaBART is trained to perform syntax-guided paraphrasing, where a source sentence containing the desired semantics and a parse tree specifying the desired syntax are given as inputs. In order to generate a paraphrase that follows the given syntax, ParaBART uses separate encoders to learn disentangled semantic and syntactic representations from their respective inputs. In this way, the disentangled representations capture sufficient semantic and syntactic information needed for paraphrase generation. The semantic encoder is also encouraged to ignore the syntax of the source sentence, as the desired syntax is already provided by the syntax input.

ParaBART achieves strong performance across unsupervised semantic textual similarity tasks. Furthermore, semantic embeddings learned by ParaBART contain significantly less syntactic information as suggested by probing results, and yield robust performance on datasets with syntactic variation. 

Our source code is available at \url{https://github.com/uclanlp/ParaBART}.

\section{Related Work}
Various sentence embedding models have been proposed in recent years. Most of these models utilize supervision from parallel data~\cite{wieting2018paranmt-50m,TACL1742,wieting2019simple,wieting2020bilingual}, natural language inference data~\cite{conneau-etal-2017-supervised,cer2018universal,reimers2019sentence-bert}, or a combination of both \cite{subramanian2018learning}. 

Many efforts towards controlled text generation have been focused on learning disentangled sentence representations \cite{pmlr-v70-hu17e,fu2018style,john2019disentangled}. In the context of disentangling semantics and syntax, \citet{bao2019generating} and \citet{chen2019multi-task} utilize variational autoencoders to learn two latent variables for semantics and syntax. In contrast, we use the outputs of a constituency parser to learn purely syntactic representations, and facilitate the usage of powerful pre-trained language models as semantic encoders.

Our approach is also related to prior work on syntax-controlled paraphrase generation \citep{iyyer2018adversarial, kumar-etal-2020-syntax, goyal2020neural, huang2021generating}. While these approaches focus on generating high-quality paraphrases that conform to the desired syntax, we are interested in how semantic and syntactic information can be disentangled and how to obtain good semantic sentence embeddings.

\section{Proposed Model -- ParaBART}

\begin{figure}[h]
\centering
\includegraphics[width=0.48\textwidth]{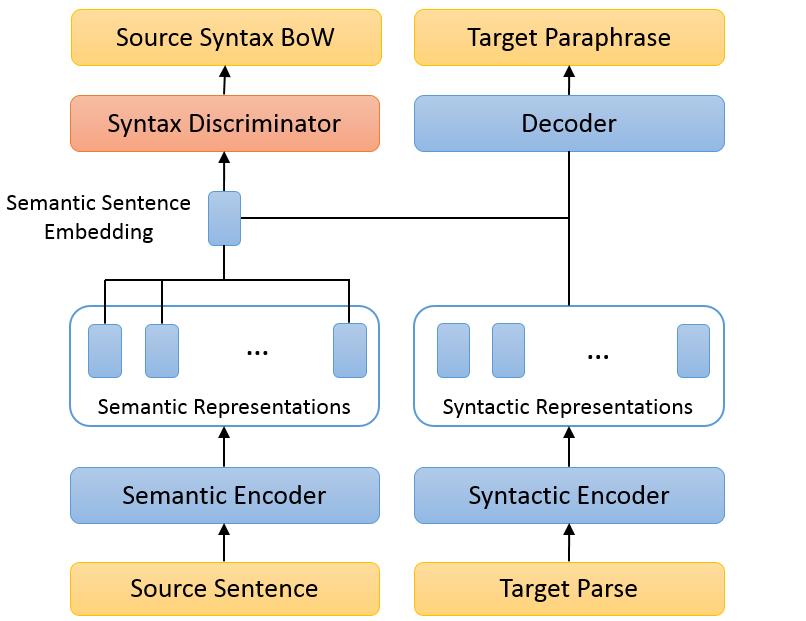}
\caption{An overview of ParaBART. The model extracts semantic and syntactic representations from a source sentence and a target parse respectively, and uses both the semantic sentence embedding and the target syntactic representations to generate the target paraphrase. ParaBART is trained in an adversarial setting, with the syntax discriminator (red) trying to decode the source syntax from the semantic embedding, and the paraphrasing model (blue) trying to fool the syntax discriminator and generate the target paraphrase at the same time.}
\label{fig:model}
\end{figure}

Our goal is to build a semantic sentence embedding model that learns to separate syntax from semantic embeddings. ParaBART is trained to generate syntax-guided paraphrases, where the model attempts to only extract the semantic part from the input sentence, and combine it with a different syntax specified by the additional syntax input in the form of a constituency parse tree. 

Figure \ref{fig:model} outlines the proposed model, which consists of a semantic encoder that learns the semantics of a source sentence, a syntactic encoder that encodes the desired syntax of a paraphrase, and a decoder that generates a corresponding paraphrase. Additionally, we add a syntax discriminator to adversarially remove syntactic information from the semantic embeddings. 

Given a source sentence $S_1$ and a target constituency parse tree $P_2$, ParaBART is trained to generate a paraphrase $S_2$ that shares the semantics of $S_1$ and conforms to the syntax specified by $P_2$. Semantics and syntax are two key aspects that determine how a sentence is generated. Our model learns purely syntactic representations from the output trees generated by a constituency parser, and extracts the semantic embedding directly from the source sentence. The syntax discriminator and the syntactic encoder are designed to remove source syntax and provide target syntax, thus encouraging the semantic encoder to only capture source semantics.

\paragraph{Semantic Encoder} 
The semantic encoder $E_{sem}$ is a Transformer encoder that embeds a sentence
$S=(s^{(1)},...,s^{(m)})$ into contextual semantic representations:
\[U=(\mathbf{u}^{(1)},...,\mathbf{u}^{(m)})=E_{sem}\left((s^{(1)},...,s^{(m)})\right).\]
Then, we take the mean of these contextual representations $\mathbf{u}^{(i)}$ to get a fixed-length semantic sentence embedding 
\[\mathbf{\bar{u}}=\frac{1}{m}\sum_{i=1}^{m}\mathbf{u}^{(i)}.\]

\paragraph{Syntactic Encoder} 
The syntactic encoder $E_{syn}$ is a Transformer encoder that takes a linearized constituency parse tree $P=(p^{(1)},...,p^{(n)})$ and converts it into contextual syntactic representations
\[V=(\mathbf{v}^{(1)},...,\mathbf{v}^{(n)})=E_{syn}\left((p^{(1)},...,p^{(n)})\right).\]
For example, the linearized parse tree of the sentence ``This book is good.'' is ``(S (NP (DT) (NN)) (VP (VBZ) (ADJP)) (.))''. Such input sequence preserves the tree structure, allowing the syntactic encoder to capture the exact syntax needed for decoding.

\paragraph{Decoder} 
The decoder $D_{dec}$ uses the semantic sentence embedding $\mathbf{\bar{u}}$ and the contextual syntactic representations $V$ to generate a paraphrase that shares semantics with the source sentence while following the syntax of the given parse tree. In other words,
\[(y^{(1)},...,y^{(l)})=D_{dec}\left(\text{Concat}(\mathbf{\bar{u}},V)\right).\]
During training, given a source sentence $S_1$, a target parse tree $P_2$ and a target paraphrase $S_2=(s_2^{1},...,s_2^{l})$, we minimize the following \emph{paraphrase generation loss}: 
\[\mathcal{L}_{para}=-\sum_{i=1}^{l}\log P(y^{(i)}=s_2^{(i)}|S_1, P_2).\]
Since the syntactic representations do not contain semantics, the semantic encoder needs to accurately capture the semantics of the source sentence for a paraphrase to be generated. Meanwhile, the full syntactic structure of the target is provided by the syntactic encoder, thus encouraging the semantic encoder to ignore the source syntax.

\paragraph{Syntax Discriminator} 
To further encourage the disentanglement of semantics and syntax, we employ a syntax discriminator to adversarially remove syntactic information from semantic embeddings. We first train the syntax discriminator to predict the syntax from its semantic embedding, and then train the semantic encoder to ``fool'' the syntax discriminator such that the source syntax cannot be predicted from the semantic embedding. 

More specifically, we adopt a simplified approach similar to \citet{john2019disentangled} by encoding source syntax as a Bag-of-Words vector $\mathbf{h}$ of its constituency parse tree. For any given source parse tree, this vector contains the count of occurrences of every constituent tag, divided by the total number of constituents in the parse tree. Given the semantic sentence embedding $\mathbf{\bar{u}}$, our linear syntax discriminator $D_{dis}$ predicts $\mathbf{h}$ by
\[\mathbf{y}_{h}=D_{dis}(\mathbf{\bar{u}})=\text{softmax}(\mathbf{W}\mathbf{\bar{u}}+\mathbf{b})\]
with the following \emph{adversarial loss}:
\[\mathcal{L}_{adv}=-\sum_{t\in T}\mathbf{h}(t)\log(\mathbf{y}_{h}(t)),\]
where $T$ denotes the set of all constituent tags.

\paragraph{Training} 
We adversarially train $E_{sem}$, $E_{syn}$, $D_{dec}$, and $D_{dis}$ with the following objective:
\[\min\limits_{E_{sem}, E_{syn}, D_{dec}} \left(\max\limits_{D_{dis}} \left(\mathcal{L}_{para}-\lambda_{adv}\mathcal{L}_{adv}\right)\right),\]
where $\lambda_{adv}$ is a hyperparameter to balance loss terms. In each iteration, we update the $D_{dis}$ by considering the inner optimization, and then update $E_{sem}$, $E_{syn}$ and $D_{dec}$ by considering the outer optimization.

\begin{table*}
\small
\centering
\begin{tabular}{l|cccccc|c}
\hline
\textbf{Model} & \textbf{STS12} & \textbf{STS13} & \textbf{STS14} & \textbf{STS15} & \textbf{STS16} & \textbf{STS-B} & \textbf{Avg.}\\
\hline
Avg. BERT embeddings \citep{devlin2019bert} & 46.9 & 52.8 & 57.2 & 63.5 & 64.5 & 47.9 & 55.5\\
Avg. BART embeddings \citep{lewis2020bart} & 50.8 & 42.8 & 56.1 & 63.9 & 59.5 & 52.0 & 54.2 \\
InferSent \citep{conneau-etal-2017-supervised} & 59.3 & 59.0 & 70.0 & 71.5 & 71.5 & 70.0 & 66.9 \\
VGVAE \citep{chen2019multi-task} & 61.8 & 62.2 & 69.2 & 72.5 & 67.8 & 74.2 & 68.0 \\
USE \citep{cer2018universal} & 61.4 & 63.5 & 70.6 & 74.3 & 73.9 & 74.2 & 69.7 \\
Sentence-BERT \citep{reimers2019sentence-bert} & 64.6 & 67.5 & 73.2 & 74.3 & 70.1 & 74.1 & 70.6 \\
BGT \citep{wieting2020bilingual} & \textbf{68.9} & \phantom{\textsuperscript{*}}62.2\textsuperscript{*} & 75.9 & 79.4 & 79.3 & - & - \\
\hline
ParaBART & 68.4 & \textbf{71.1} & \textbf{76.4} & 80.7 & \textbf{80.1} & 78.5 & \textbf{75.9} \\
 - w/o adversarial loss & 67.5 & 70.0 & 75.8 & \textbf{80.9} & 80.0 & \textbf{78.7} & 75.5 \\
 - w/o adversarial loss and syntactic guidance & 66.4 & 65.3 & 73.6 & 80.0 & 78.6 & 75.4 & 73.2 \\
\hline
\end{tabular}
\caption{Pearson's $r$  (in percentage) between cosine similarity of sentence embeddings and gold labels on STS tasks from 2012 to 2016 and STS Benchmark test set. BGT results are taken from \citet{wieting2020bilingual}. *BGT is evaluated on an additional dataset from STS13, which is not included in the standard SentEval toolkit.} 
\label{tab:sts}
\vspace{-1em}
\end{table*}

\section{Experiments}
In this section, we demonstrate that ParaBART is capable of learning semantic sentence embeddings that capture semantic similarity, contain less syntactic information, and yield robust performance against syntactic variation on semantic tasks.
\subsection{Setup}
We sample 1 million English paraphrase pairs from ParaNMT-50M \citep{wieting2018paranmt-50m}, and split this dataset into 5,000 pairs as the validation set and the rest as our training set.
The constituency parse trees of all sentences are obtained from Stanford CoreNLP \citep{manning-etal-2014-stanford}. We fine-tune a 6-layer BART\textsubscript{base} encoder as the semantic encoder and the first BART\textsubscript{base} decoder layer as the decoder for our model. 

We train ParaBART on a GTX 1080Ti GPU using AdamW \citep{loshchilov2018decoupled} optimizer with a learning rate of $2\times10^{-5}$ for the encoder and syntax discriminator, and $1\times10^{-4}$ for the rest of the model. The batch size is set to 64. All models are trained for 10 epochs, which takes about 2 days to complete. The maximum length of input sentences and linearized parse trees are set to 40 and 160 respectively. We set the weight of adversarial loss to 0.1. Appendix \ref{apx:impl} shows more implementation details.

\paragraph{Baselines}
 We compare our model with other sentence embeddings models, including InferSent \citep{conneau-etal-2017-supervised}, Universal Sentence Encoder (USE) \citep{cer2018universal}, Sentence-BERT\textsubscript{base} \citep{reimers2019sentence-bert}, VGVAE \citep{chen2019multi-task}, and BGT \citep{wieting2020bilingual}. We also include mean-pooled BERT\textsubscript{base} and BART\textsubscript{base} embeddings. In addition to ParaBART, we consider two model ablations: ParaBART without adversarial loss, and ParaBART without syntactic guidance and adversarial loss.

\subsection{Semantic Textual Similarity}
We evaluate our semantic sentence embeddings on the unsupervised Semantic Textual Similarity (STS) tasks from SemEval 2012 to 2016 (\citealp{agirre-etal-2012-semeval}; \citeyear{agirre-etal-2013-sem}; \citeyear{agirre-etal-2014-semeval}; \citeyear{agirre-etal-2015-semeval}; \citeyear{agirre-etal-2016-semeval})  and STS Benchmark test set \citep{cer2017semeval-2017}, where the goal is to predict a continuous-valued score between 0 and 5 indicating how similar the meanings of a sentence pair are. For all models, we compute the cosine similarity of embedding vectors as the semantic similarity measure. We use the standard SentEval toolkit \citep{conneau2018senteval} for evaluation and report average Pearson correlation over all domains. 

As shown in Table \ref{tab:sts}, both average BERT embeddings and average BART embeddings perform poorly on STS tasks, as the entanglement of semantic and syntactic information leads to low correlation with semantic similarity. Training ParaBART on paraphrase data substantially improves the correlation. With the addition of syntactic guidance and adversarial loss, ParaBART achieves the best overall performance across STS tasks, showing the effectiveness of our approach.

\begin{table}
\small
\centering
\begin{tabular}{l|ccc}
\hline
\text{Model} & \textbf{BShift} & \textbf{TreeDepth} & \textbf{TopConst}\\
\hline
Avg. BART embed. & 90.5 & 47.8 & 80.1 \\
ParaBART & \textbf{72.4} & \textbf{33.9} & \textbf{67.2} \\ 
 -  w/o AL  & 75.4 & 36.6 & 71.7 \\
 -  w/o AL and SG & 83.3 & 46.5 & 83.1 \\

\hline
\end{tabular}
\caption{Results on syntactic probing tasks. Semantic embeddings with lower accuracy on downstream syntactic tasks contain less syntactic information, suggesting better disentanglement of semantics and syntax. AL and SG denote adversarial loss and syntactic guidance, respectively.}
\label{tab:sg}
\vspace{-0.5em}
\end{table}

\subsection{Syntactic Probing}

To better understand how well our model learns to disentangle syntactic information from semantic embeddings, we probe our semantic sentence embeddings with downstream syntactic tasks. Following \citet{conneau2018what}, we investigate to what degree our semantic sentence embeddings can be used to identify bigram word reordering (BShift), estimate parse tree depth (TreeDepth), and predict parse tree top-level constituents (TopConst). Top-level constituents are defined as the group of constituency parse tree nodes immediately below the sentence (S) node. We use the datasets provided by SentEval \citep{conneau2018senteval} to train a Multi-Layer Perceptron classifier with a single 50-neuron hidden layer on top of semantic sentence embeddings, and report accuracy on all tasks. 

As shown in Table \ref{tab:sg}, sentence embeddings pooled from pre-trained BART model contain rich syntactic information that can be used to accurately predict syntactic properties including word order and top-level constituents. The disentanglement induced by ParaBART is evident, lowering the accuracy of downstream syntactic tasks by more than 10 points compared to pre-trained BART embeddings and ParaBART without adversarial loss and syntactic guidance. The results suggest that the semantic sentence embeddings learned by ParaBART indeed contain less syntactic information.

\begin{table}[t]
\small
\centering
\begin{tabular}{l}
\hline
\textbf{QQP-Easy} \\
\hline
What are the essential skills of the project management? \\
What are the essential skills of a project manager? \\
\hline
\textbf{QQP-Hard} \\
\hline
Is there a reason why we should travel alone? \\
What are some reasons to travel alone?\\
\hline
\end{tabular}
\caption{Examples of paraphrase pairs from \textit{QQP-Easy} and \textit{QQP-Hard}.}
\label{tab:eg}
\vspace{-1em}
\end{table} 

\subsection{Robustness Against Syntactic Variation}
Intuitively, semantic sentence embedding models that learn to disentangle semantics and syntax are expected to yield more robust performance on datasets with high syntactic variation. We consider the task of paraphrase detection on Quora Question Pairs \citep{qqp} dev set as a testbed for evaluating model robustness. We categorize paraphrase pairs based on whether they share the same top-level constituents. We randomly sample 1,000 paraphrase pairs from each of the two classes, combined with a common set of 1,000 randomly sampled non-paraphrase pairs, to create two datasets \textit{QQP-Easy} and \textit{QQP-Hard}. Paraphrase pairs from \textit{QQP-Hard} are generally harder to identify as they are much more syntactically different compared to those from \textit{QQP-Easy}. Table \ref{tab:eg} shows some examples from these two datasets. We evaluate semantic sentence embeddings on these datasets in an unsupervised manner by computing the cosine similarity as the semantic similarity measure. We search for the best threshold between -1 and 1 with a step size of 0.01 on each dataset, and report the highest accuracy. The results are shown in Table \ref{tab:qqp}.

While Universal Sentence Encoder scores much higher than other models on \textit{QQP-Easy}, its performance degrades significantly on \textit{QQP-Hard}. In comparison, ParaBART demonstrates better robustness against syntactic variation, and surpasses USE to become the best model on the more syntactically diverse \textit{QQP-Hard}. It is worth mentioning that even pre-trained BART embeddings give decent results on \textit{QQP-Easy}, suggesting large overlaps between paraphrase pairs from \textit{QQP-Easy}. On the other hand, the poor performance of pre-trained BART embeddings on a more syntactically diverse dataset like \textit{QQP-Hard} clearly shows its incompetence as semantic sentence embeddings.

\begin{table}
\small
\centering
\begin{tabular}{l|cc}
\hline
\textbf{Model} & \textbf{QQP-Easy} & \textbf{QQP-Hard} \\
\hline
Avg. BART embed. & 72.3 & 64.1 \\
InferSent & 72.1 & 67.5 \\
VGVAE & 71.5 & 67.1 \\
USE & \textbf{80.7} & 72.4 \\
Sentence-BERT & 74.3 & 70.7 \\
\hline
ParaBART & 76.5 & \textbf{72.7} \\
 - w/o AL & 76.8 & 72.1 \\
 - w/o AL and SG & 76.1 & 69.9 \\
\hline
\end{tabular}

\caption{Results on \textit{QQP-Easy} and \textit{QQP-Hard}. For every model we report the highest accuracy after finding the best threshold. AL and SG denote adversarial loss and syntactic guidance, respectively.}
\label{tab:qqp}
\vspace{-0.5em}
\end{table}

\section{Conclusion}
In this paper, we present ParaBART, a semantic sentence embedding model that learns to disentangle semantics and syntax in sentence embeddings from pre-trained language models. Experiments show that our semantic sentence embeddings yield strong performance on unsupervised semantic similarity tasks. Further investigation demonstrates the effectiveness of disentanglement, and robustness of our semantic sentence embeddings against syntactic variation on downstream semantic tasks.

\section*{Acknowledgments}
We thank anonymous reviewers for their helpful feedback. We thank UCLA-NLP group for the valuable discussions and comments. This work is supported in part by Amazon Research Award. 

\section*{Ethics Considerations}
Our sentence embeddings can potentially capture bias reflective of the training data we use, which is a common problem for models trained on large annotated datasets. While the focus of our work is to disentangle semantics and syntax, our model can potentially generate offensive or biased content learned from training data if it is used for paraphrase generation. We suggest carefully examining the potential bias exhibited in our models before deploying them in any real-world applications.

\bibliography{nlp, new}
\bibliographystyle{acl_natbib}

\clearpage

\appendix
\section{Implementation Details}
\label{apx:impl}

\paragraph{Datasets}
We use the ParaNMT-50M dataset released by \citet{wieting2018paranmt-50m}, which can be obtained from \url{https://github.com/jwieting/para-nmt-50m}. We sample 1 million English paraphrase pairs from ParaNMT-50M, and split this dataset into 5000 pairs as the validation set and the rest as our training set. STS and syntactic probing datasets are directly taken from SentEval, which can be accessed from \url{https://github.com/facebookresearch/SentEval}. Quora Question Pairs are downloaded from the official GLUE Benchmark website (\url{https://gluebenchmark.com/}).

\paragraph{Word Dropout} 
We observe that some paraphrase pairs in our training set contain many overlapping words, which means our model can learn to generate the target paraphrase by just copying words from a source sentence without fully understanding the semantics of the sentence. To alleviate this issue, we apply word dropout \citep{Iyyer:Manjunatha:Boyd-Graber:III} that randomly masks a portion of the input tokens. We don't apply word dropout to syntactic inputs, as these inputs are designed to provide the exact syntactic structure of the paraphrase and encourage disentanglement of syntactic and semantic representations. We set the word dropout probability to 0.2 for all our models.

\paragraph{Hyperparameter Search}
Hyperparameters of ParaBART are tuned manually based on the paraphrase generation loss on the validation set. Specifically, the weight of adversarial loss is tuned within \{0.1, 0.2, 0.5, 1.0\}. Word dropout is selected from \{0.0, 0.1, 0.2, 0.4\}. Learning rate is tuned within \{1,2,5,10\}$\times10^{-5}$.

None of the previous models we compare in this work involves any hyperparameter search. The results for BGT are taken from \citet{wieting2020bilingual}. For all other sentence embedding models, we use the trained model provided by their respective authors. These models include InferSent (\url{https://github.com/facebookresearch/InferSent}, USE (\url{https://tfhub.dev/google/universal-sentence-encoder-large/2}), Sentence-BERT\textsubscript{base} (\url{https://github.com/UKPLab/sentence-transformers}) and VGVAE (\url{https://github.com/mingdachen/syntactic-template-generation}).

Performance on STS and QQP are evaluated under unsupervised settings. For syntactic probing tasks that involve training classifiers, we report the accuracy on the validation set provided by SentEval in Table \ref{tab:val}.

\begin{table}[h]
\small
\centering
\begin{tabular}{l|ccc}
\hline
\text{Model} & \textbf{BShift} & \textbf{TreeDepth} & \textbf{TopConst}\\
\hline
Avg. BART embed. & 90.4 & 47.5 & 80.2 \\
ParaBART & 73.0 & 34.8 & 67.6 \\ 
 -  w/o AL  & 75.4 & 36.7 & 72.1 \\
 -  w/o AL and SG & 84.0 & 46.7 & 82.7 \\

\hline
\end{tabular}
\caption{Validation accuracy on syntactic probing tasks.}
\label{tab:val}
\end{table}



\end{document}